# On ρ in a Decision-Theoretic Apparatus of Dempster-Shafer Theory

**Johan Schubert**
*Division of Information System Technology,
Department of Command and Control Warfare Technology,
National Defence Research Establishment, Stockholm, Sweden.*

**ABSTRACT**

*Thomas M. Strat has developed a decision-theoretic apparatus for Dempster-Shafer theory (Decision analysis using belief functions,* Intern. J. Approx. Reason. *4(5/6), 391-417, 1990). In this apparatus, expected utility intervals are constructed for different choices. The choice with the highest expected utility is preferable to others. However, to find the preferred choice when the expected utility interval of one choice is included in that of another, it is necessary to interpolate a discerning point in the intervals. This is done by the parameter ρ, defined as the probability that the ambiguity about the utility of every nonsingleton focal element will turn out as favorable as possible. If there are several different decision makers, we might sometimes be more interested in having the highest expected utility among the decision makers rather than only trying to maximize our own expected utility regardless of choices made by other decision makers. The preference of each choice is then determined by the probability of yielding the highest expected utility. This probability is equal to the maximal interval length of ρ under which an alternative is preferred. We must here take into account not only the choices already made by other decision makers but also the rational choices we can assume to be made by later decision makers. In Strats apparatus, an assumption, unwarranted by the evidence at hand, has to be made about the value of ρ. We demonstrate that no such assumption is necessary. It is sufficient to assume a uniform probability distribution for ρ to be able to discern the most preferable choice. We discuss when this approach is justifiable.*

## 1. INTRODUCTION

To make rational decisions under uncertainty is somewhat complicated in Dempster-Shafer theory (Dempster [1], Shafer [2]) because of the interval representation. In [3] Nguyen and Walker discussed different approaches to decision making with belief functions. They found three different basic models. The first is based on the Choquet integral that yields the expected utility with respect to belief functions;

$$E_F(u) = \int_0^\infty F(u>t)dt + \int_{-\infty}^0 [F(u>t)-1]\,dt$$

where $F$ is a belief function defined on $2^\Theta$ by $F(A) = inf\{P(A): P \in \mathbf{P}\}$ and $\mathbf{P} = \{P: F \leq P\}$ is a class of probability measures on $\Theta$. This leads to the pessimistic strategy of ranking alternatives by their minimal expected utility.

In the second basic model the decision maker uses some additional information or subjective views. Instead of searching for the alternative that maximizes expected utility, the utility function will be supplemented by some new function dependent on the utility and some other parameter corresponding to the additional information or subjective views. An article by Strat [4] is an example of the second basic model.

The third basic model consists of models using the insufficient reason principle or equivalently the maximum entropy principle. As an example, Smets and Kennes [5] have developed a two-level model of credal belief and pignistic probability, called the *transferable belief model* (TBM).

On the credal level of this model the reasoning process takes place in the usual manner as within Dempster-Shafer theory. Here beliefs are held by belief functions and combined by Dempster's rule. When a decision must be taken, the belief on the credal level is transformed to a probability at the pignistic level by a *pignistic transformation* based on Laplace's insufficient-reason principle;

$$BetP(x) = \sum_{x \subseteq A \in \Re} \frac{m(A)}{|A|} = \sum_{A \in \Re} m(A) \cdot \frac{|x \cap A|}{|A|},$$
$$BetP(B) = \sum_{A \in \Re} m(A) \cdot \frac{|B \cap A|}{|A|}$$

where $BetP(\cdot)$ is the pignistic probability we should use to 'bet' with in a utility maximization process. Here $\Re$ is the set of all propositions. It is called the betting frame.

The pignistic probability regarding some proposition $A$ depends on the organization of the betting frame $\Re$. But regardless of the organization of the betting frame we always have $BetP(A) \geq Bel(A) \quad \forall A \in \Re$.

Further discussions on decision making with belief functions can be found in [6, 7].

This article is concerned with a method that has recently been developed by Strat [4]. In this method an expected utility interval is constructed for each choice;

$[E_*(x), E^*(x)]$

where $E_*(\cdot)$ and $E^*(\cdot)$ are defined as

$$E_*(x) \triangleq \sum_{A_i \subseteq \Theta} inf(A_i) \cdot m_\Theta(A_i)$$

and

$$E^*(x) \triangleq \sum_{A_i \subseteq \Theta} sup(A_i) \cdot m_\Theta(A_i),$$

$\Theta$ is a frame of discernment, i.e., an exhaustive set of mutually exclusive possibilities, and $m_\Theta$ is a basic probability assignment (bpa), a function from the power set of $\Theta$ to $[0, 1]$:

$$m_\Theta: 2^\Theta \to [0,1]$$

whenever

$$m_\Theta(\emptyset) = 0$$

and

$$\sum_{A_i \subseteq \Theta} m_\Theta(A_i) = 1.$$

The frame of discernment is here the set of all possible utilities of the outcomes. We will call $E_*$ the lower expected utility and $E^*$ the upper expected utility.

Our preference among different alternatives will depend upon their expected utility. Let the expected utility be defined as

$$E(x) \triangleq E_*(x) + \rho \cdot [E^*(x) - E_*(x)]$$

where $\rho$ is defined as the probability that the ambiguity about the utility of every nonsingleton focal element will turn out as favorably as possible, i.e. the probability that nature will turn out as favorably as possibly towards us as decision makers.

Strat gives an example involving a carnival wheel (Figure 1). This wheel is divided into 10 equal sectors, each one having a payoff of either $1, $5, $10, or $20. One of the sectors is hidden from view. He asks: How much are we willing to pay to play this game?

The frame of discernment $\Theta$ is {$1, $5, $10, $20}. Assume that

$$m(\{\$1\}) = 0.4,$$
$$m(\{\$5\}) = 0.2,$$
$$m(\{\$10\}) = 0.2,$$
$$m(\{\$20\}) = 0.1,$$
$$m(\{\$1, \$5, \$10, \$20\}) = 0.1.$$

Calculating the expected value interval we have

$E(x) = [E_*(x), E^*(x)]$
$= [0.4 \times \$1 + 0.2 \times \$5 + 0.2 \times \$10 + 0.1 \times \$20 + 0.1 \times \$1,$
$0.4 \times \$1 + 0.2 \times \$5 + 0.2 \times \$10 + 0.1 \times \$20 + 0.1 \times \$20]$
$= [\$5.50, \$7.40].$

Thus, we would be willing to pay at least $5.50, but certainly not more than $7.40. But should we be willing to play for $6?

Obviously, when we are searching for the most preferable choice we can immediately disregard those choices where the upper expected utility is less than the highest lower expected utility among all choices. Furthermore, if both interval limits of the utility interval are higher for one alternative than for another, i.e. $E_{i*} > E_{j*}$ and $E_i^* > E_j^*$, then the first one, choice $i$, is always preferable regardless of the value of $\rho$. In fact, if we receive the choices in decreasing order of the magnitude of their upper expected utility we can immediately disregard any choice whose lower expected utility is less than any lower expected utility of the previous choices. Only if the expected-utility interval of one choice is included in the interval of another choice will our preference depend on the assumed value of $\rho$. As a result, we will end up with a set of expected utility intervals ordered by interval inclusion, $[E_{1*},E_1^*] \subseteq [E_{2*},E_2^*] \subseteq \ldots \subseteq [E_{n*},E_n^*] \subseteq [0,1]$. Here we have renumbered the choices by the order of interval inclusion, i.e. the order of increasing interval length. In the following we will only consider choices ordered and renumbered by interval inclusion.

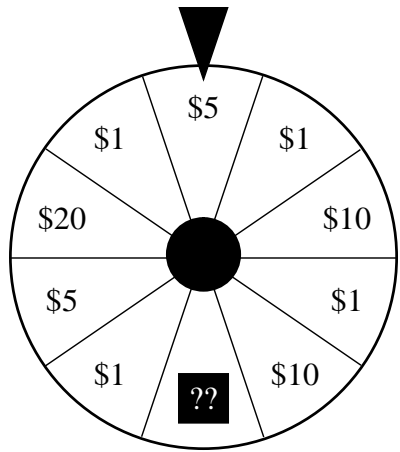

**Figure 1.** Carnival wheel.

Strat argues that instead of first assuming a value for $\rho$ and then calculating the choice that results, one might ask the reverse question. At what value of $\rho$ would I be willing to change my decision?

Let us study the choice between $x_1$ and $x_2$ where $[E_{1*},E_1^*] \subseteq [E_{2*},E_2^*]$;

*choice* 1: $[E_{1*},E_1^*]$,

*choice* 2: $[E_{2*},E_2^*]$.

Here choice 1 is preferred when

$$E_{1*} + \rho \cdot (E_1^* - E_{1*}) > E_{2*} + \rho \cdot (E_2^* - E_{2*})$$

We find that the two choices are equally preferable if

$$\rho = \frac{E_{1*} - E_{2*}}{(E_2^* - E_{2*}) - (E_1^* - E_{1*})}.$$

Let us call this value $\rho_{12}$. Since choice 1 has the higher lower expected utility of the two choices, it is preferred when $\rho \in [0,\rho_{12}]$, and choice 2 is preferred when $\rho \in [\rho_{12},1]$.

Let choice 1 be the decision not to play, and choice 2 the decision to play:

*choice* 1: $E(x) = [\$6.00, \$6.00]$,

*choice* 2: $E(x) = [\$5.50, \$7.40]$.

When we compare these two choices, we find that choice 1 is preferable to choice 2 when

$$\rho_{12} < \frac{E_{1*} - E_{2*}}{(E_2^* - E_{2*}) - (E_1^* - E_{1*})} = \frac{6.00 - 5.50}{(7.40 - 5.50) - (6.00 - 6.00)}$$
$$= \frac{0.50}{1.90} = 0.263.$$

Thus, when $\rho < 0.263$ we will not play the game, but when $\rho > 0.263$ we will.

Furthermore, this might best be seen as an indication of whether we need to gather more information or are ready to make a decision. If $\rho$ is close to 0 or 1, we might be willing to make a decision right now, but if, on the other hand, it is around 0.5, we would prefer to gather additional information before making a decision.

However, at times we might be forced to make a decision right now regardless of the value of $\rho$. Rather than trying to estimate $\rho$ in this situation, we might choose a different route.

In this article we will establish an alternative to making an outright, and often unwarranted, assumption about $\rho$. This alternative is to accept a uniform probability distribution for $\rho$.

Adopting a uniform probability distribution for $\rho$ requires two conditions being fulfilled. Firstly, there certainly must not be any evidence at hand regarding the value of the probability $\rho$. Such evidence could, for example, be in the form of domain knowledge, direct evidence regarding the value of $\rho$ or knowledge that the decision situation is controlled by either the decision maker or an adversary. It would seem to be commonplace that there is no direct evidence available regarding the value of $\rho$. The situation we are looking for is then a business like situation in a field with poor domain knowledge where the outcomes are not controlled by either the decision maker or an adversary, i.e. a decision situation without evidence regarding the value of $\rho$. Secondly, it must be a decision situation where the decision maker is not only interested in minimizing the expected loss regardless of the possible gains or interested in maximizing the expected gain regardless of the possible losses. In these two situations he would choose to adopt $\rho = 0$ or $\rho = 1$, respectively, even if there is no available evidence regarding the value of $\rho$. This would be the situation if the decision maker were forced to play a game he thinks is unfavorable. Then he would try to minimize the expected loss, i.e. choose $\rho = 0$. If, on the other hand, the decision maker is forced to obtain a lot of value by playing a particular game, he may try to maximize the expected gain, i.e. choose $\rho = 1$. This eliminates the extreme situations where the decision maker is forced into a game by one reason or another, i.e. situations where it is not possible to avoid a choice. What is remaining is the "normal" businesslike decision situations where we do not have a reason to choose one value for ρ over another: when there is not any evidence at hand regarding the value of ρ.

As Strat points out in his article, if we make an assumption about the value of ρ we should not confuse our assumption about ambiguity with our risk preference. Our risk preference is handled by adopting utilities.

The methodology in this article was developed as the decision part of a multiple-target tracking algorithm (Schubert [8], Bergsten and Schubert [9]) for an antisubmarine intelligence analysis system.

In Section 2 we will discuss points of preference change and in Section 3 the uniform probability distribution for ρ. In Section 4 we will study decision making with a uniform probability distribution for ρ, and the different objectives decision makers might have when there are several decision makers competing. Finally, conclusions are drawn in Section 5.

## 2. THE PREFERRED CHOICE

When we have several choices they may be preferred in different intervals of ρ. If we calculate all $\rho_{ij}$'s and order them by increasing magnitude we can calculate the expected utility of every choice for a point in each interval of the ordered $\rho_{ij}$'s. The choice with the highest expected utility in each interval is then the preferred choice for that interval. However, we already know that choice 1 is preferred when ρ = 0, since this choice has the highest lower expected utility among all choices, and it will remain the preferred choice while ρ is less than $\min_i \rho_{1i}$, the smallest of all $\rho_{ij}$'s and the first point of preference change. Beyond this point, choice $i$ will be preferable over choice 1. Since choice 1 will never again be preferred in any other interval, we may now disregard all other $\rho_{1j}$, $j \neq i$, even though they represent points of possible preference change. The reason for this is obvious, choice 1 can never again be the most preferable choice for any interval above $\min_i \rho_{1i}$ since it is not even preferred to choice $i$ beyond that point. Thus, these points of possible preference change will never represent an actual change of the current preference. Continuing, choice $i$ will now be preferred up to the point where $\rho = \min_j \rho_{ij}$, and beyond this point choice $j$ will be preferred up until $\rho = \min_k \rho_{jk}$, etc. Thus, by iteration we find that the choices are each preferable in the following intervals:

*choice* 1: [0, $min_i \rho_{1i}$],

*choice i*: [$min_i \rho_{1i}$, $min_j \rho_{ij}$],

*choice j*: [$min_j \rho_{ij}$, $min_k \rho_{jk}$],

⋮

*choice n*: [$\rho_{mn}$, 1].

Alternatively, for any choice $j$ that is preferable somewhere, its interval of preference can be described as

*choice j*: $[max_i\ \rho_{ij},\ min_k\ \rho_{jk}]$.

If two or more $\rho_{ij}$'s are equal in a minimization, $min_j\ \rho_{ij}$, the next preferred choice will be ambiguous. In this case we take the choice with the highest number. If not, we would end up with one or more choices preferred under a zero interval length of $\rho$ before we would get this choice anyway.

## 3. A UNIFORM PROBABILITY DISTRIBUTION FOR $\rho$

All we know about the value of $\rho$ is that it is a parameter the belongs to the set of real numbers between 0 and 1, $\rho \in [0, 1]$, i.e. we know that a frame of all possible values of $\rho$ is that same set of numbers, $\Theta = [0, 1]$. Thus, apart from knowing the frame for $\rho$ we do not know anything at all. We have a vacuous bpa where $m(\Theta) = 1$. In order not to reduce the overall nonspecificity of this initial state when making an assumption about the probability distribution about $\rho$, we might ask that any such assumption about $\rho$ should yield the same nonspecificity as what we have now. We define the nonspecificity as

$$I(m) = \sum_{A \in F} m(A) \cdot Log_2|A|,$$

which is a generalization of Hartley's information [10].

Calculating the nonspecificity $I(m)$ of this initial state where $F = \{\Theta\}$ and $m(\Theta) = 1$, we have

$$I(m) = \sum_{A \in F} m(A) \cdot Log_2|A|$$
$$= m(\Theta) \cdot Log_2|\Theta| = 1 \cdot Log_2|\Theta|,$$

and since $\Theta$ is the infinite set of real numbers between 0 and 1, we obtain an infinite nonspecificity.

If we make a single-point assumption about $\rho$ where $F = \{\{\rho\}\}$ and $m(\{\rho\}) = 1$, we obtain a nonspecificity of

$$I(m) = \sum_{A \in F} m(A) \cdot Log_2|A|$$
$$= m(\{\rho\}) \cdot Log_2|\{\rho\}| = 1 \cdot Log_2 1 = 0,$$

and for any pointwise distribution for ρ where $F = \{\{\rho_1\}, \{\rho_2\}, ...\}$ we get

$$I(m) = \sum_{A \in F} m(A) \cdot Log_2|A|$$
$$= \sum_{A \in F} m(A) \cdot Log_2 1 = 0.$$

Obviously, our distribution needs a continuous part to reach the infinite nonspecificity of the initial state. Any such distribution with just one continuous part, $B$, will reach infinite nonspecificity. We have

$$I(m) = \sum_{A \in F} m(A) \cdot Log_2|A|$$
$$= m(B) \cdot Log_2|B|,$$

where $F = \{B, \{\rho_1\}, \{\rho_2\}, ...\}$ and $B$ is an interval of real numbers included in $[0, 1]$. If $B$ is of infinite size we have an infinite nonspecificity.

Furthermore, we might also demand that the nonspecificity of our new distribution should be equal to the original assignment for any size of the frame. Let $F = \{B_1, B_2, ..., \{\rho_1\}, \{\rho_2\}, ...\}$, where the $B_i$'s are intervals included in $[0, 1]$. We must then have

$$Log_2|\Theta| = \sum_{A \in F} m(A) \cdot Log_2|A|.$$

Here $A \subseteq \Theta$, and thus we may write $|A| = \alpha_A|\Theta|$, where

$$\frac{1}{|\Theta|} \leq \alpha_A \leq 1$$

and

$$\alpha_{\{\rho_i\}} = \frac{1}{|\Theta|}.$$

We have

$$Log_2|\Theta| = \sum_{A \in F} m(A) \cdot Log_2(\alpha_A \cdot |\Theta|) = \sum_{A \in F} m(A) \cdot Log_2\alpha_A + \sum_{A \in F} m(A) \cdot Log_2|\Theta|$$
$$= \sum_{A \in F} m(A) \cdot Log_2\alpha_A + Log_2|\Theta| \cdot \sum_{A \in F} m(A).$$

From this it immediately follows that

$$\sum_{A \in F} m(A) \cdot Log_2\alpha_A = 0.$$

Since $m(A) > 0$ for every $A$ and $\log_2 \alpha_A \leq 0$ for every $\alpha_A$, we must have that $\alpha_A = 1$ for every $A$. But since $|A| = \alpha_A |\Theta|$ and $\Theta$ is the entire frame, it follows that $A = \Theta$, i.e., that we have only one focal element $F = \{\Theta\}$.

This means that we have only one continuous part of the probability distribution for ρ, and that it covers the entire interval from 0 to 1, i.e. a uniform probability distribution.

## 4. DECISION MAKING

### 4.1. Decision Making With a Uniform Probability Distribution for ρ

If we refrain from making an unwarranted assumption about the value of ρ, we might instead accept a uniform probability distribution for ρ, i.e. the assumption that all values of ρ are equally probable. Any of the above choices that are preferable somewhere might now be preferred. However, the probabilities for the choices to be preferred are not equal. This probability varies with the length of the interval over which it is preferred.

If we are only interested in simple maximizing of utility then adopting a uniform probability distribution for ρ yields the same result as setting ρ = 0.5. Then, for simplicity, we might as well set ρ = 0.5 and choose the alternative that yields the highest expected utility as our decision.

However, in a situation with several different decision makers, we might sometimes be more interested in having the highest expected utility among the decision makers rather than only trying to maximize our own expected utility. Thus, rather than actually making a random assumption about ρ in order to find a preferable choice, it makes sense to prefer the choice that would most likely be preferred *if* the value of ρ were determined at random. Assuming the uniform probability distribution for ρ, this is obviously the choice that is preferred under the maximum interval length of ρ. This might be according to the principle "it is better to choose what is most likely the best alternative rather than to gamble for it." The interval length under which a choice is preferred, Pref(·), is here defined as

$$\text{Pref}(x_j) \triangleq max(0, min_k \; \rho_{jk} - max_i \; \rho_{ij})$$

where $min_k \; \rho_{nk} \triangleq 1$ and $max_i \; \rho_{i1} \triangleq 0$.

If the number of alternatives is equal to the number of decision makers, then all we have to do is to choose the alternative that is preferred under the maximal interval length. That will be the choice with the highest probability of giving us the highest expected utility.

The situation becomes more complex when the number of decision makers is less than the number of choices.

**4.2. An Example**

Let us consider an example with four choices whose expected utility intervals are ordered by interval inclusion:

*choice* 1: [0.5, 0.6],
*choice* 2: [0.4, 0.7],
*choice* 3: [0.3, 0.9],
*choice* 4: [0.2, 1.0].

Calculating the points of preference change gives us

$$\rho_{12} = \frac{E_{1*} - E_{2*}}{(E_2^* - E_{2*}) - (E_1^* - E_{1*})} = \frac{0.5 - 0.4}{(0.7 - 0.4) - (0.6 - 0.5)} = 0.5,$$

and by the same formula $\rho_{13} = 0.4$, $\rho_{14} = 0.43$, $\rho_{23} = 0.33$, $\rho_{24} = 0.4$, $\rho_{34} = 0.5$. We find by iteration that the choices are preferable in the following intervals of $\rho$:

*choice* 1: [0, $min_i\ \rho_{1i}$] = [0, $\rho_{13}$] = [0, 0.4],
*choice* 3: [0.4, $min_j\ \rho_{3j}$] = [0.4, $\rho_{34}$] = [0.4, 0.5],
*choice* 4: [0.5, 1],

and are preferred for the following interval lengths:

$$Pref(x_1) = 0.4,$$

$$Pref(x_2) = 0,$$

$$Pref(x_3) = 0.1,$$

$$Pref(x_4) = 0.5.$$

In this case choice 2 will never be preferred, regardless of the value of $\rho$. If an unwarranted assumption is made about $\rho$, any of the other three choices could be preferred. If, on the other hand, we only assume a uniform probability distribution for $\rho$, choice 4 will be considered preferable, since it is preferred for the maximum interval length of $\rho$.

**4.3. An Algorithm for Finding the Preferred Choice**

We may now find the preferred choice given a uniform probability distribution by the following algorithm.

ALGORITHM    *Let S be the empty set.*
  1. *Order and renumber all choices by falling magnitude of upper expected utility.*
  2. *For i = 1 to n do*
     2.1. *Add all choices whose expected utility interval belongs to the set of intervals ordered by interval inclusion; if $E_{i*} > E_{i-1*}$ then $S := S + \{[E_{i*}, E^*_i]\}$.*
  3. *Renumber all choices in S in order of increasing interval length magnitude.*
  4. *For all combinations of pairs in S calculate*

$$\rho_{ij} = \frac{E_{i*} - E_{j*}}{(E^*_j - E_{j*}) - (E^*_i - E_{i*})}.$$

  5. $\rho_c := 0$, $i := 1$, *maximum_preference := 0.*
  6. *Calculate the intervals of preference for each choice and find the most preferable choice;*
     *While $i \neq n$ do*
     6.1. $\rho'_c := min_j \rho_{ij}$, *where* $min_k \rho_{nk} \stackrel{\Delta}{=} 1$.
     6.2. *Pref($x_i$) = $\rho'_c$ - $\rho_c$.*
     6.3. *If Pref($x_i$) > maximum_preference then*
        6.3.1. *maximum_preference := Pref($x_i$), preferred_choice := i.*
     6.4. *i := j.*
     6.5. $\rho_c := \rho'_c$
  7. *Answer preferred_choice.*

### 4.4. Possible Refinements

Instead of changing from the strongest possible assumption of a point value for ρ to the weakest possible assumption of a uniform probability distribution, we may occasionally have a reason to assume some other probability distribution for ρ (Figure 2).

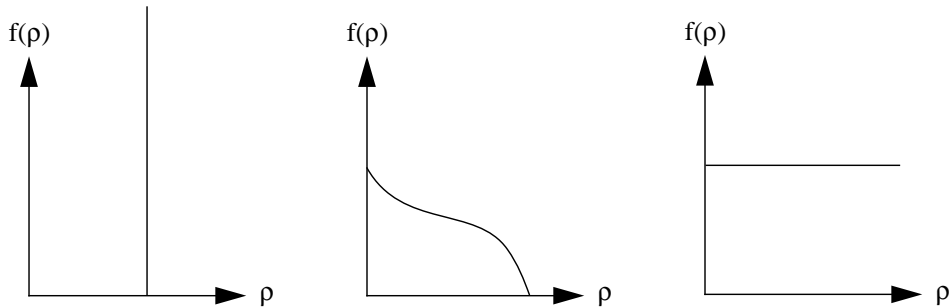

**Figure 2.** A point-valued, arbitrary and uniform probability distribution for ρ.

We might for instance have some knowledge regarding a lower and upper bound for ρ. Let us call these bounds the lower ambiguity probability $\rho_*$ and the upper ambiguity probability $\rho^*$, respectively. These bounds force a simple change in the definition of preference, Pref(·):

$$\text{Pref}(x_j) \triangleq max(0, min(\rho^*, min_k\ \rho_{jk}) - max(\rho_*, max_i\ \rho_{ij}))$$

where $min_k\ \rho_{nk} \triangleq 1$ and $max_i\ \rho_{i1} \triangleq 0$,

To incorporate the new definition of preference into the algorithm we make the following change in step 6.2.,

6.2. $Pref(x_i) = max(0, min(\rho^*, \rho'_c) - max(\rho_*, \rho_c))$,

giving all choices preferred in intervals outside the bounds of lower and upper ambiguity probability a preference of zero.

Obviously, we must be able to assume any probability distribution $f(\rho)$ for ρ. We can make a general definition of preference as

$$Pref(x_j) \triangleq max(0, \int_{max_i\ \rho_{ij}}^{min_k\ \rho_{jk}} f(\rho)d\rho)$$

where $min_k\ \rho_{nk} \triangleq 1$ and $max_i\ \rho_{i1} \triangleq 0$.

Finally, we change the computation of preference in step 6.2 of the algorithm to

$$6.2.\ \ Pref(x_j) \triangleq F(\rho'_c) - F(\rho_c)$$

where

$$F(\rho) = \int f(\rho)d\rho.$$

### 4.5. Two Decision Makers Searching for the Most Preferable Choice

When two decision makers compete for the highest utility, the preference of each alternative is determined by the chance of having the alternative that is preferred for the maximal interval length of ρ after our opponent has also made his choice. If we assume we have the first choice, then our opponent will make his choice taking into account the choice we made. Since our goal is to have the highest possible probability of having the best alternative, we must also take into account the best choice our opponent can make. It is found by choosing the alternative with the highest preference as defined by

$$\text{Pref}(x_j) \triangleq min(min_k\ \rho_{jk}, 1 - max_i\ \rho_{ij}).$$

Here $\min_k \rho_{jk}$ is the preference for choice $j$ when our opponent chooses his best alternative $k$ where $k > j$ and $1 - \max_i \rho_{ij}$ is the preference for choice $j$ when our opponent chooses his best alternative $i$ where $i < j$.

If, on the other hand, we are the second of the two decision makers, the situation is even simpler. We just have to find the choice with maximal preference as defined by

$$\text{Pref}(x_j, k) = \begin{cases} \rho_{jk}, j < k \\ \rho_{kj}, j > k \end{cases}$$

where $k$ is the alternative already chosen by our opponent.

### 4.6. Several Decision Makers

When the number of decision makers is less than the number of choices, the situation becomes much more complex. We must here take into account not only the choices already made be other decision makers, but also the rational choices we can assume to be made by later decision makers. This is because the length of the preference interval for any alternative depends on the other choices that are made. If $I^*$ is the set of all choices made by previous decision makers, the preference of a choice $x_j$ may be calculated as

$$\text{Pref}(x_j, I^*) = max\left(0, \min_{k \in I^* + I_*(I^*, j)} \rho_{jk} - \max_{i \in I^* + I_*(I^*, j)} \rho_{ij}\right),$$

where $I_*(I^*, j)$ is the set of rational choices the later decision makers will make given our choice $j$. For any decision maker we want to find the alternative that maximizes his preference, i.e.

$$\max_{j \in I - I^*} \text{Pref}(x_j, I^*)$$

where $I$ is the set of all possible choices.

This problem is solved starting with the final choice made by the last of the $n$ decision makers, and for all possible sets of earlier choices $I^*$. Here $|I^*| = n - 1$ and $I_* = \varnothing$. We find the earlier choices by stepping backwards through all possible sequences of choices made by different decision makers until we reach the first choice made by the first decision maker.

This can be seen as going "up" a tree with one decision maker at each level until we reach the first decision maker at the root of the tree. Each branch at a certain level of the tree corresponds to a different sequence of choices made by the earlier decision makers. The edges going "down" from each node at this level corresponds to the possible choices that can be made by the decision maker at this level.

## 5. CONCLUSION

We have demonstrated that it is not necessary to make a point-value assumption about ρ in Strat's decision-theoretic apparatus of Dempster-Shafer theory. In fact, it is sufficient to assume a uniform probability distribution for ρ to be able to discern the most preferable choice. We have given an algorithm for finding the most preferable choice based on an iterative search of points of preference change among choices ordered by interval inclusion. We discuss the ability to assume any probability distribution for ρ.

We also discussed the more complex problem of several decision makers competing for the highest expected utility. The preference for each alternative to some decision makers was shown to be the probability that the alternative has the highest expected utility after all decision makers have made their choices, where we take into account both the choices already made be other decision makers and the rational choices we can assume to be made by later decision makers.

## ACKNOWLEDGMENTS

I would like to thank Stefan Arnborg, Ulla Bergsten, and Per Svensson for their helpful comments regarding this article.